\documentclass[11pt,a4paper]{article}
\usepackage[hyperref]{ranlp2021}
\usepackage{times}
\usepackage{latexsym}
\usepackage{tabularx}

\usepackage{microtype}
\usepackage{graphicx}
\graphicspath{{./final_images/}}
\aclfinalcopy 

\newcommand*{\affaddr}[1]{#1} 
\newcommand*{\affmark}[1][*]{\textsuperscript{#1}}
\newcommand*{\email}[1]{\texttt{#1}}

\title{Opinion Prediction with User Fingerprinting}

\author{
Kishore Tumarada\affmark[1], Yifan Zhang\affmark[1], Fan Yang\affmark[2],\\ \textbf{Eduard Dragut\affmark[3], Omprakash Gnawali\affmark[1], and Arjun Mukherjee\affmark[1]}\\
\affaddr{\affmark[1]Department of Computer science, University of Houston, USA}\\
\affaddr{\affmark[2]Amazon, USA \{The paper was done before the author joined Amazon.\}}\\
\affaddr{\affmark[3]Department of Computer and information systems, Temple University, USA}\\
\email{ \{kishore.t04, aeryen, fyangskip\}@gmail.com,}\\ \email{edragut@temple.edu, \{odgnawal, amukher6\}@central.uh.edu }

}

\date{}

\begin{document}
\maketitle
\begin{abstract}
Opinion prediction is an emerging research area with diverse real-world applications, such as market research and situational awareness. We identify two lines of approaches to the problem of opinion prediction. One uses topic-based sentiment analysis with time-series modeling, while the other uses static embedding of text. The latter approaches seek user-specific solutions by generating user fingerprints. Such approaches are useful in predicting user's reactions to unseen content. In this work, we propose a novel dynamic fingerprinting method that leverages contextual embedding of user's comments conditioned on relevant user's reading history. We integrate BERT variants with a recurrent neural network to generate predictions. The results show up to 13\% improvement in micro F1-score compared to previous approaches. Experimental results show novel insights that were previously unknown such as better predictions for an increase in dynamic history length, the impact of the nature of the article on performance, thereby laying the foundation for further research.
\end{abstract}

\section{Introduction}
Sentiment analysis plays a key role in economic, social, and political contexts. Companies can understand customer's opinions based on reviews and/or social media conversations to make fast and accurate product decisions. They can index unstructured customer data at scale based on broad sentiments such as positive, negative, or neutral. It also enables market research - explore new markets, anticipate future trends, and seeking an edge on the competition. In a political context, sentiment analysis can be useful in understanding political homophily using tweets analysis \citep{DBLP:journals/jisa/CaetanoLSM18}. 

Moreover, as social media has emerged as a new source of communication, governments can analyze people's reactions to issues such as police-related encounters, mob protests and anticipate responses before they turn violent. In this context, predicting people's opinion before expressing is an important next step in applying sentiment analysis to real-world applications. We are approaching this problem through sentiment prediction, especially on current news events, that can raise situational awareness, understand the future viewpoints of the citizenry on pressing social and political issues.

Recently fine-grained models generated a person's fingerprint, based on one's recent reading and response history, to predict response on an unknown event  \citep{DBLP:conf/coling/YangDM20}. In this work, we propose a novel architecture to create a dynamic fingerprint of a user that is contingent upon the target event. We choose to evaluate our models on the dataset used by \citealp{DBLP:conf/coling/YangDM20}, which contains newspaper articles as events and user's comments on them as opinions. Those ground truth opinions are used as a basis for sentiment prediction in this work.

Our model consists of three main steps. In the first step, relevant articles are extracted from a user's reading history based on their similarity to target article.In the second step, the contextual embedding of these relevant articles, conditioned on the target article, is used to create a reading track. Similarly, we create a writing (response) track with the contextual embedding of extracted articles with corresponding comments by the user. Lastly, dynamic fingerprints are generated based on the temporal pattern of the reading and writing tracks. These dynamic fingerprint vectors for a particular user are then used to predict the user's sentiment on the target event.

Our main contributions in this paper are:
\begin{enumerate}
\item A novel architecture of dynamic fingerprint generation based on the contextual embedding of the user's reading history.

\item Experimental results show that our method outperformed the previous approach over various news outlets datasets.
\end{enumerate}

\section{Related work}

Opinion prediction based on the temporal pattern of sentiments is a relatively new research topic, but basic concepts such as sentiment analysis, analysis of user comments, question answering based on dialogue context have been explored in different communities and settings.

\citealp{DBLP:conf/acl/SiMLLLD13} proposed a technique to leverage topic-based Twitter sentiments to predict the stock market using vector autoregression and Dirichlet process mixture models. \citealp{DBLP:journals/access/LiWZZ19} proposed a time+user dual attention-based LSTM network to perform emotional analysis on Chinese public opinion texts on social networking platform.\citealp{DBLP:conf/coling/YangMZ16} leveraged multiple domains for sentiment classification by learning high-level features that are able to generalize across domains. But they did not use contextual embedding and explore the prospect of generating a unique fingerprint of users before predicting sentiment.

\citealp{DBLP:journals/widm/HeHMOD20} studied the commenting activity on news articles and analysed the differences of patterns compared to social media comments. \citealp{DBLP:journals/pvldb/LiuDMM15} proposed a FLORIN system that provides support for near real-time applications such as sentiment analysis of user comments on daily news. \citealp{DBLP:conf/icwsm/HeSMVD21} studied the problem of predicting total number of user comments on a news article based on features such as topics, early dynamics of user comments, news factor among others. But they did not perform sentiment prediction of users based on their past comments.

Conversational question answering (CQA) is an emerging research area in the machine reading comprehension task (MRC). For single-turn MRC tasks, contextualized language representation using BERT has obtained state-of-the-art scores on SQuAD datasets  \citep{DBLP:conf/naacl/DevlinCLT19}. CQA is a multi-turn question answering task that includes passage comprehension, contextual understanding, and coreference resolution. Zhu et al. have proposed SDNet \citep{DBLP:journals/corr/abs-1812-03593} to solve this problem by concatenating previous questions and answers as one query to fuse context into traditional MRC models by leveraging BERT, attention, and RNN techniques. Similarly, \citealp{DBLP:journals/corr/abs-1905-12848} have proposed fine-tuning approach with BERT in a multi-turn context by modeling the interaction between paragraph and dialogue history independently.

However, these models cannot be applied to the present problem since they did not integrate the concepts of the sequential pattern of sentiments along with the unique fingerprint of each user, which can play a key role in predicting the future opinion of a user on different topics.

\begin{figure*}[t]
    \centering
    \includegraphics[width=1.0\textwidth]{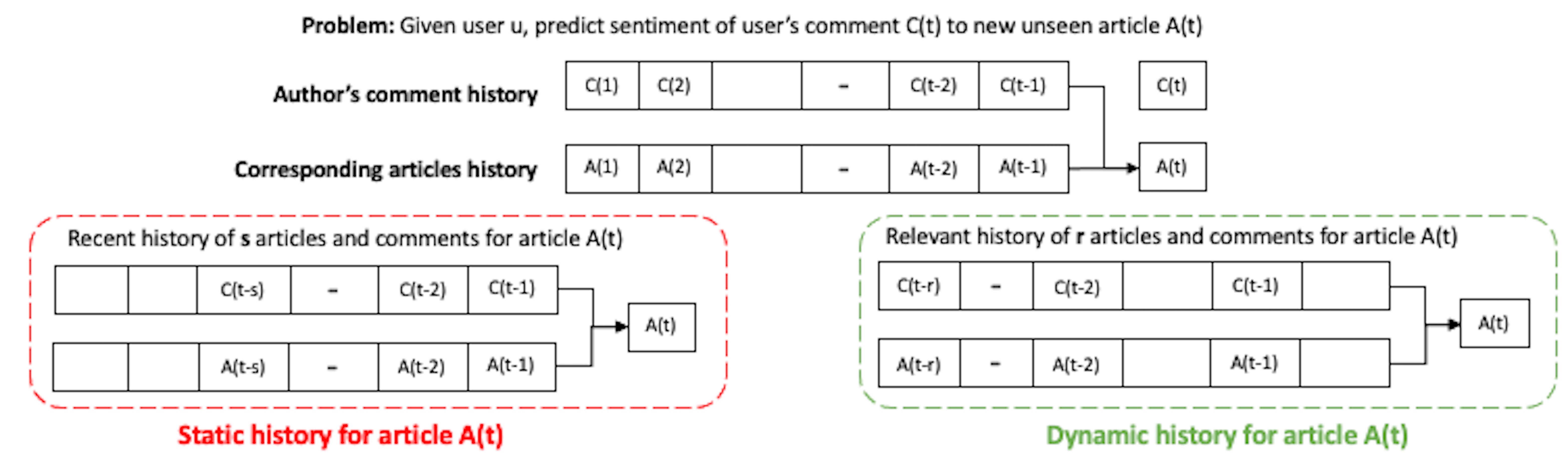}
    \caption{History selection techniques for dynamic and static fingerprints. In dynamic fingerprint, for a target article A(t) at time t, $r$ relevant articles (in green) from user's history A(0) to A(t-1) (in white) are selected based on semantic textual similarity between A(t) and history. In static fingerprint, $s$ articles (in red) from t-s to t-1 time steps in immediate history are selected.}
    \label{fig:history_sel}
\end{figure*}

\begin{table*}[t]

\textbf{Target article: Pelosi Says House Will Condemn All Hate as Anti-Semitism Debate Overshadows Congress}

\begin{tabularx}{\textwidth}{|c|X|X|}

 \hline
 \textbf{S No.} &
 \textbf{Static history} 
 &
 \textbf{Dynamic history}\\
 \hline
 
 \centering 1 & Ilhan Omar Knows Exactly What She Is Doing & With Control of Congress at Stake, Trump Reprises a Favorite Theme: Fear Immigrants\\
 \hline
 2 & Ilhan Omar Controversy Caps a Month of Stumbles for Democratic Leaders	& After Loss of House, Trump Makes Overture to Democrats, Coupled With Threats\\
 \hline
 3 & Tom Brokaw: What Trump and Nixon Share	& Pompeo Speech Lays Out Vision for Mideast, Taking Shots at Obama\\
 \hline
 4 & Pelosi Says House Will Condemn All Hate as Anti-Semitism Debate Overshadows Congress	& White House Considers Using Storm Aid Funds as a Way to Pay for the Border Wall\\
 \hline
 5 & Tariff Man Has Become Deficit Man	& Senate Leaders Plan Competing Bills to End Shutdown\\
 \hline
 6 & Paul Manafort to Be Sentenced Thursday in 1 of 2 Cases Against Him	& House Votes to Block Trump’s National Emergency Declaration About the Border\\
 \hline
 
 \end{tabularx}
 \caption{Articles extracted by different methods of history selection. We can see that dynamic history of articles are more relevant than static history.}

\label{table:article_hist_samples}
\end{table*}
\section{Proposed model and methodology}\label{sectionmodel}

In this section, we propose two classes of \textbf{F}inger\textbf{P}rint \textbf{E}mbedding models  (FPE) - Static and Dynamic - for the task of predicting the sentiment of a user $u$ to a new article. In a narrow sense, we used the term \textbf{static} to refer to the approach of using recent comment history, which is independent of the nature of target article A(t), and \textbf{dynamic} to refer to the approach of using articles relevant to target article A(t) in the overall history of user's comments, which are dependent on the nature of target article. In a broad sense, static and dynamic terms distinguish the way target article A(t) is integrated with user's reading history to generate the fingerprint.

We used the user's commenting history on articles that they read. We assume that we know the sentiment of each comment. (This can be obtained with one of the many sentiment analysis tools.) Formally, we are given the articles, comments along with the sentiments of a user $u$, i.e., $(A_1, C_1, S_1), (A_1, C_2, S_2)$,..., $(A_2, C_j, S_j)$,  ..., $(A_{t-1}, C_n, S_n)$, and the goal is to predict the sentiment $S_t$ of $u$'s response to unseen article $A_t$. In general, $n > t$ because a user may post multiple comments to an article.

The overall architecture includes history selection, text embedding, fingerprint creation, and lastly sentiment prediction. We describe them below.

\begin{figure*}[t]
    \centering
    \includegraphics[width=1.0\textwidth]{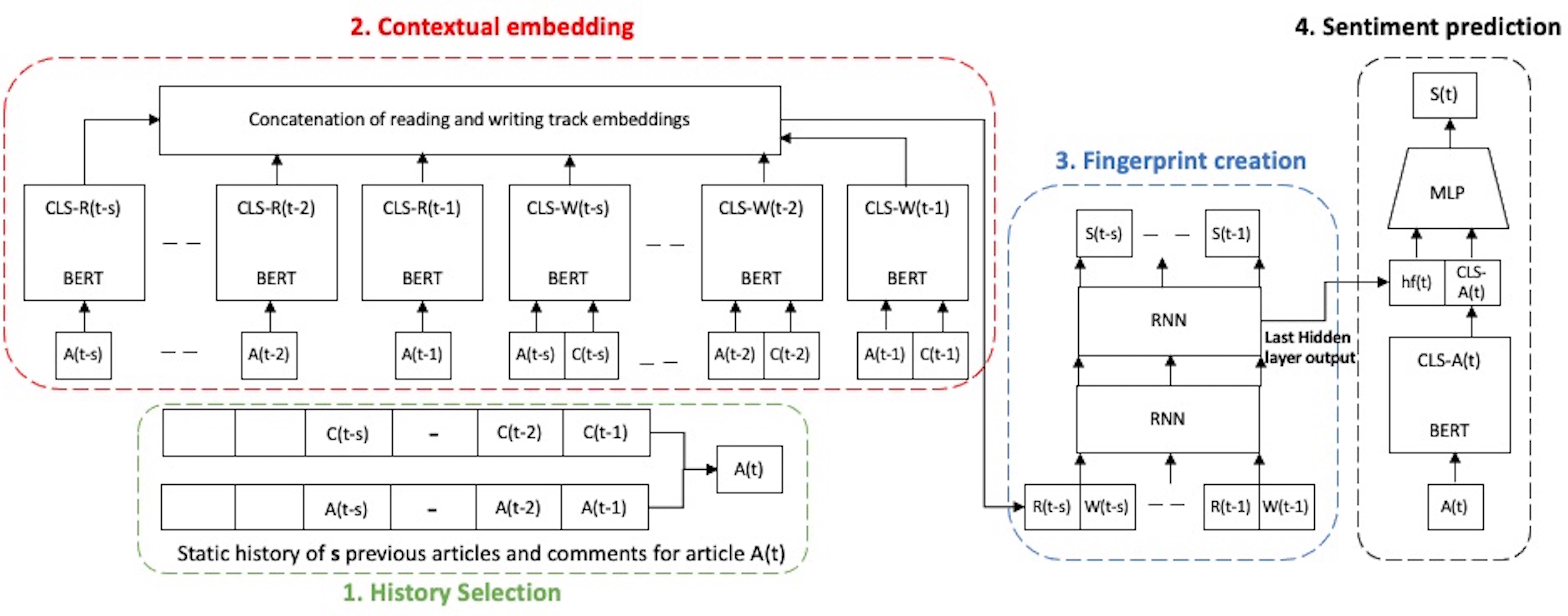}
    \caption{For the Static FPE model, in stage 1, for a target article A(t), $s$ article-comment pairs from (t-s) to (t-1) immediate past from a particular user's history are selected. In stage 2, the contextual embedding of $s$ articles are generated, and corresponding $[CLS]$ tokens from DistilBERT/ELECTRA models are extracted as reading track R(t-s) to R(t-1). Similarly, contextual embedding of articles A(t-s) to A(t-1) conditioned with corresponding comments C(t-s) to C(t-1) are generated and $[CLS]$ tokens are extracted as writing track W(t-s) to W(t-1). In stage 3, an RNN with 2 hidden layers is trained with concatenation of both tracks against corresponding comment sentiment S(t) at each time step t and the last hidden layer output is extracted as the fingerprint of user. In the last stage, a Multi-layer perceptron (MLP) is trained on the concatenation of the fingerprint $h^{f}_t$ and $CLS^{A}_t$ against the sentiment S(t) of the user's comment on article A(t).}
    \label{fig:staticmodel}
\end{figure*}

\subsection{History selection}
We have explored two methods of history selection - static history and dynamic history. 

Given an article $A_t$, its static history, according to the user $u$, is the list of the most recent article-comment pairs posted by $u$. Depending on the magnitude of $s$ and number of comments made by the user, the list may include comments from one article or multiple articles. We use this method in the Static FPE model.

For the dynamic history of article $A_t$, we have ranked all articles of author's reading history based on similarity and picked top $r$ articles, along with their comments, as shown in Figure \ref{fig:history_sel}. Here similarity between articles is calculated using DistilRoBERTa Semantic textual similarity model \citep{DBLP:conf/emnlp/ReimersG19}. 


It is a variant of sentence-BERT, a modified pre-trained BERT network that uses siamese and triplet network structures to derive semantically meaningful sentence embeddings that are then compared using the cosine-similarity metric.

Table \ref{table:article_hist_samples} shows the article samples extracted by both methods. For the target article "Pelosi says house will condemn all hate as anti-semitism debate overshadows congress", we can see that articles in the dynamic history, especially articles 1,3,4 are more related to the themes such as congress, hate etc., of target article. But static history does not show any relevant articles, except for article 4, which was also because the user has read the target article in the recent past. So we can see that dynamic history has a more pertinent set of articles than static history.  

\subsection{Text embedding}

In this stage, we create reading and writing tracks based on the selected user's history. Specifically, we encode articles and comments using contextual embedding by two types of encoders (BERT variants) - DistilBERT \citep{DBLP:journals/corr/abs-1910-01108} and ELECTRA \citep{DBLP:journals/corr/abs-2003-10555} models. However, the proposed models are open to any variant of BERT encoder. We have used encoders in two modes - single-sentence mode where single span of contiguous text is encoded in form a special classification token $[CLS]$; two-sentence mode where two spans of text separated by $[SEP]$ token are encoded as $[CLS]$. 

In the static FPE model, as shown in Figure \ref{fig:staticmodel}, reading track comprises of selected articles, encoded by fixed-length $[CLS]$ tokens of single-sentence mode of BERT variant models, i.e., $[CLS^R_{t-s},.., CLS^R_{t-2}, CLS^R_{t-1}]$. We have conducted experiments both on - pretrained (with frozen parameters) and trained - BERT variants. Writing track comprises of both article and comments at each time step k $[A_k, C_{i_k}]$, encoded as $[CLS]$ token outputs of two-sentence mode of BERT variant models, similar to reading track, i.e., $[CLS^W_{t-s},.., CLS^W_{t-2}, CLS^W_{t-1}]$ . 

\begin{figure*}[t]
    \includegraphics[width=\textwidth]{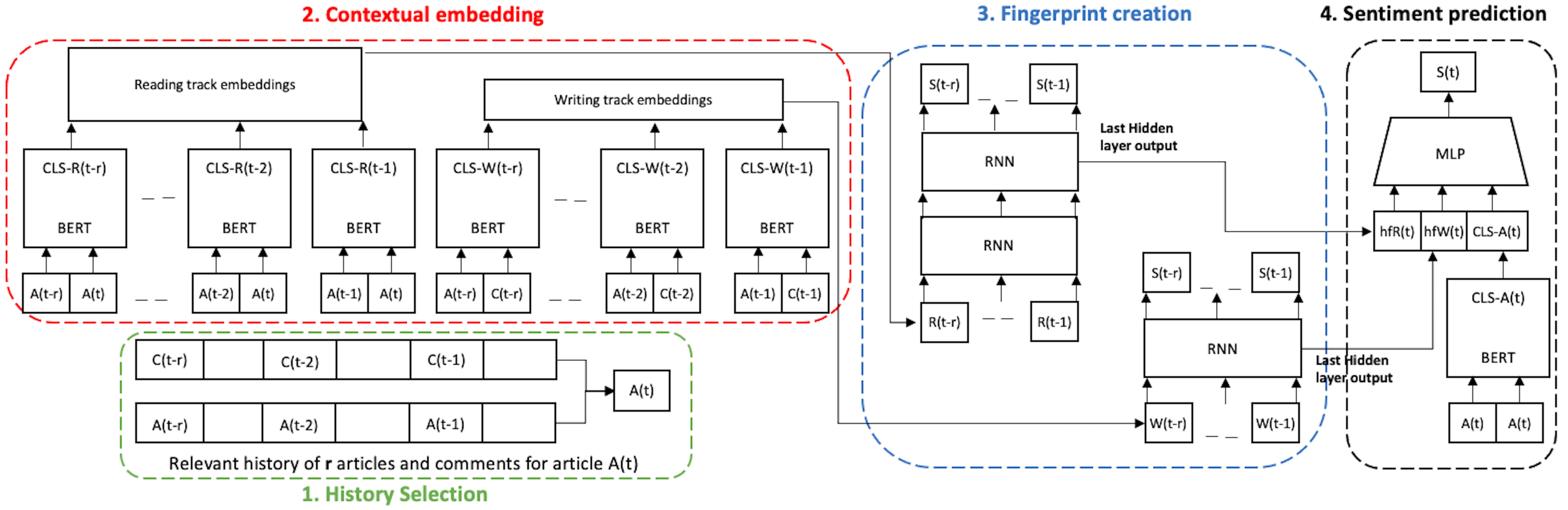}
    \caption{For the Dynamic FPE model, in stage 1, for a target article A(t), $r$ relevant article-comment pairs from a particular user's complete history are selected based on semantic textual similarity. In stage 2, the contextual embedding of each of the $r$ articles conditioned with target article A(t) is generated and corresponding $[CLS]$ tokens from DistilBERT/ELECTRA models are extracted as reading track. Similarly, the contextual embedding of articles A(t-r) to A(t-1) conditioned with corresponding comments C(t-r) to C(t-1) are generated, and $[CLS]$ tokens from DistilBERT/ELECTRA models are extracted as writing track. In stage 3, the reading track is encoded with a two-layered RNN trained against the corresponding comment sentiment S(t-r) to S(t-1) at each time step and the last hidden layer output is extracted as the reading fingerprint of the user. Similarly, the writing track is also encoded with a one-layered RNN and the last hidden layer output is extracted as a writing fingerprint. In the last stage, an MLP is trained on the concatenation of the reading fingerprint $h^{f_{R}}_t$, writing fingerprint $h^{f_{W}}_t$ and $CLS^{A}_t$ against sentiment S(t) of corresponding comment C(t) of user on article A(t).}
    \label{fig:dynamicmodel}
\end{figure*}

In the dynamic FPE model, as shown in Figure \ref{fig:dynamicmodel}, for the reading track - target article $A_t$ is appended to every article of relevant history and both of them are encoded by two-sentence mode of BERT variant models as $[CLS]$ token outputs, i.e., $[CLS^R_{t-r},.., CLS^R_{t-2}, CLS^R_{t-1}]$. The writing track is similar to that of static FPE model, i.e., $[CLS^W_{t-r},.., CLS^W_{t-2}, CLS^W_{t-1}]$.

Lastly, both the tracks are concatenated at each time step to create a unified fingerprint in case of static FPE model. But in dynamic FPE model, these tracks are used as separate entities to create two different fingerprints in the next stage.

\subsection{Fingerprint creation}

In this stage, unidirectional RNN is used to form contextual understanding of both reading and writing tracks. 

In the static FPE model, each $[CLS]$ token of reading track is concatenated with corresponding $[CLS]$ token of writing track at each time step, i.e., $[ {CLS^{R}}_{t-s}, {CLS^{W}}_{t-s};$  ... ${CLS^{R}}_{t-2}, {CLS^{W}}_{t-2};$  ${CLS^{R}}_{t-1}, {CLS^{W}}_{t-1}; ]$. We have used Gated Recurrent Unit (GRU)  \citep{DBLP:journals/corr/ChungGCB14} instead of LSTM, because the former has fewer parameters, trains faster with comparable performance to the latter. A two-layer GRU network is trained with the above concatenated output against corresponding sentiments $S_{t-s}$ ...,  $S_{t-2}, S_{t-1}$. The last hidden layer output is taken as the fingerprint embedding $h^{f}_t$ for article $A_t$ of user $u$. 

In the Dynamic FPE model, we create a separate fingerprint for the reading track and writing track, respectively, with 2-layer RNN and 1-layer RNN networks. Both are GRU networks. An additional layer is used for the reading track to get a more complex feature representation of the relationship between target articles and the relevant history of articles. Here, fingerprint embedding is the concatenation of the last hidden layer outputs of both networks, i.e., $[ h^{f_{R}}_t ; h^{f_{W}}_t ]$. 

\subsection{Sentiment prediction}

Lastly, a one-layer Multi-Layer Perceptron (MLP) is trained with concatenation of fingerprint embedding and $[CLS]$ token embedding of the target article, as input against the sentiment of response. In the static FPE model, the MLP is trained on $[ h^{f}_t; CLS^{A}_t]$ against output $S_{t}$. Whereas in the dynamic FPE model, it is trained on  $[ h^{f_{R}}_t ; h^{f_{W}}_t;  CLS^{A}_t]$ against output $S_{t}$.

\section{Experiments}

The main goals of our experiments are: 
\begin{enumerate}
\item Measure the performance (in terms of micro F1-score) of both static and dynamic model variants in the prediction of the sentiment of a user to an unknown article

\item Analysis of model performance by studying the impact of dynamic history length and nature of articles on prediction.
\end{enumerate}

\subsection{Data preparation}

We perform our empirical study on the datasets used for Personal opinion prediction by \citealp{DBLP:conf/coling/YangDM20}.In these datasets, news articles are randomly selected from Archiveis, The Guardian, and New York Times. We do not consider users with fewer than ten comments. If after this step an article remains without any user, the article is discarded. We checked manually a random subset of articles and their comments and found that irrelevant comments are very few to ignore. 

Each input example case comprises a target article/comment and its corresponding selected history of article-comment pairs. For each user, the data is split into training and test sets in the ratio of 90:10 ratio sequentially, i.e., the last 10\% of comments made by user chronologically as test data and the remaining as training data. Also, the training set is split into training and validation data in the ratio of 90:10 preserving the sequential order. 

\begin{figure*}[t]
    \centering 
    \includegraphics[width=\textwidth]{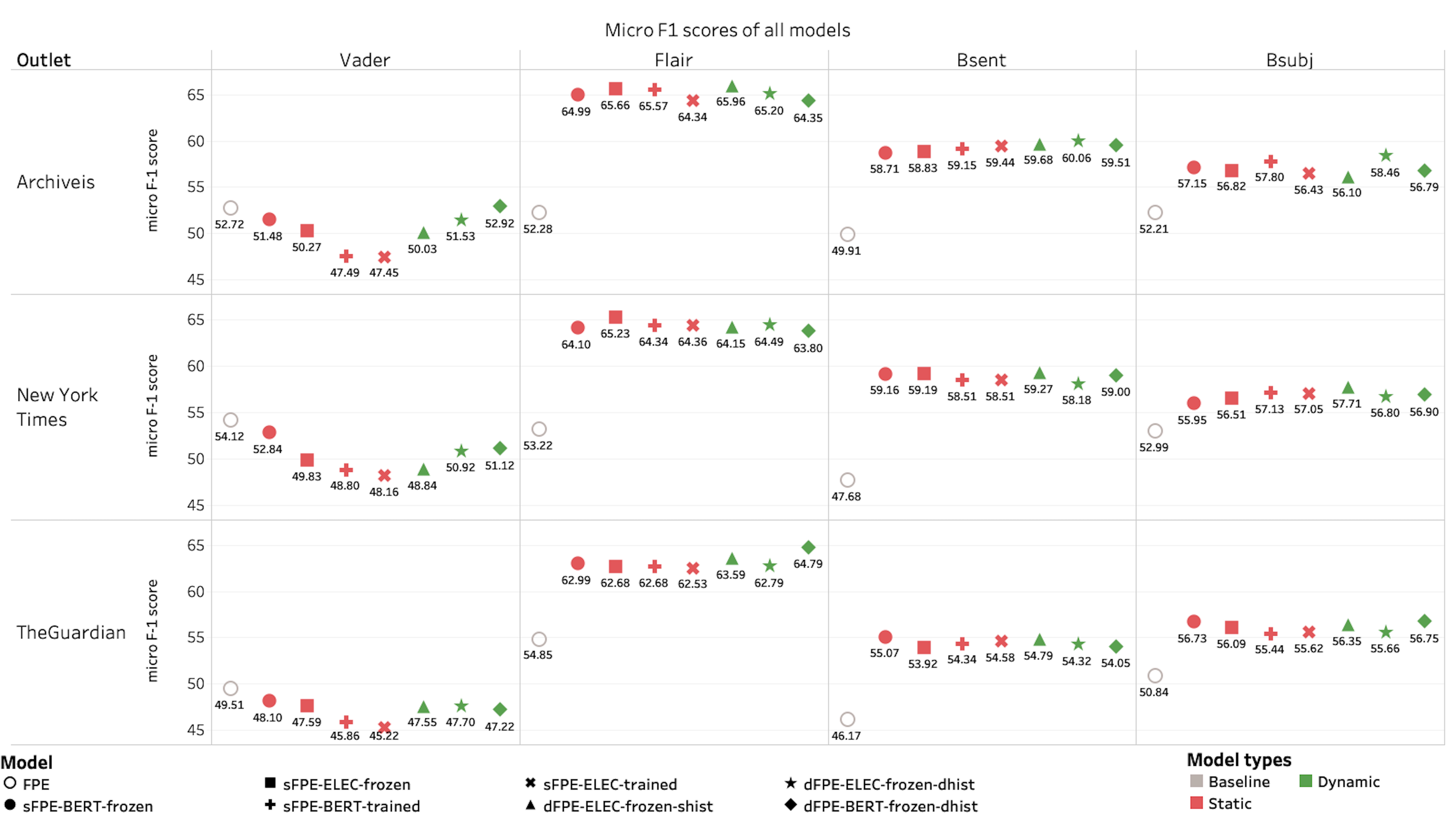}
    \caption{Micro F1-scores of all models on test data for all outlets with baseline scores, shown as white empty circle in the figure. We can see that dynamic FPE models have better performance of 4-13\% points over baseline FPE model, except for Vader sentiment, across all three outlets.}
    \label{fig:final_results}
\end{figure*}

Since our task is to predict the sentiment (as a score in [-1, 1]) on a future comment, we considered 4 models to conceive the sentiment score. They are - Vader \citep{DBLP:conf/icwsm/HuttoG14}, Flair \citep{DBLP:conf/naacl/AkbikBBRSV19}, BlobText sentiment, and BlobText subjectivity (Loria et al., 2014), to automatically label all comments. We assume that users have consistent views and stances on the same event within these articles and comments. 
\textbf{Vader} is a rule-based model for general sentiment analysis. It is constructed from a generalizable, valence-based, human-curated gold standard sentiment lexicon. When assessing the sentiment of tweets, Vader outperforms individual human raters \citep{DBLP:conf/icwsm/HuttoG14}. \textbf{Flair} presents a unified interface for all word embeddings and supports methods for producing vector representation of entire documents. We use the Flair pre-trained classification model for sentiment labels. The model is trained on the IMDB dataset and has 90.54 micro F1-score. \textbf{BlobText} is a simple rule-based API for sentiment analysis. It has both sentiment model and subjectivity model, and we refer to them as Bsent and Bsubj respectively.

We used article titles and comment content as the basis for our model. We did not use article content since they are extremely long and moreover our focus is on user's opinion on the article, not the article per se. Table \ref{table:dataset} shows the dataset statistics. 

\begin{table}[t]
\centering

\small
 \begin{tabularx}{0.5\textwidth}{Xlll}

 \hline
 Statistics &Archiveis &TheGuardian &NewYorkTimes\\
 \hline
 \# $U$ &20,920 	&41,069 	&37,957\\
 \hline
 \# Mean $C_{u}$ &25.8	&33.63	&32.26\\
 \hline
 \# Med. $C_{u}$ &9	&13	&10\\
 \hline
 \# $A$ &2,043	&6,393	&3,647 \\
 \hline
 \# $C$ &812,768	&5,467,755	&2,328,597 \\
 \hline

 \end{tabularx}

 \caption{Statistics for three news datasets. For each dataset, \# $U$ refers to total number of users, \# mean $C_u$ - mean number of comments per user, \# med. $C_u$ - median number of comments per user, \# $A$ - total number of articles and \# $C$ - total number of comments.}
\label{table:dataset}
\end{table}

\subsection{Baselines and experiment settings}

We evaluated our models against the baseline Fingerprint embedding (FPE) model \citep{DBLP:conf/coling/YangDM20}. In FPE, recent history of a target article was extracted and then Byte-Pair Embedding (BPE) \citep{DBLP:conf/lrec/Heinzerling018} and GRU were used to encode the words in articles and comments into fixed-length vectors. Subsequently, user's fingerprint was generated using a second GRU that modeled the sequence of history, which was a direct concatenation of prior articles and comments, encoded as fixed-length vectors. Finally, the concatenation of fingerprint embedding and target article embedding was given to MLP to predict the sentiment.

On the contrary, we examined both recent and relevant history of target article. We also used BERT based contextual embedding to encode the relationship between articles and comments rather than separate encoding. Finally, we created fingerprint separately for reading and writing track and then concatenated in the final stage for predicting the sentiment.


\begin{figure*}[t]
    \centering
    \includegraphics[width=\textwidth]{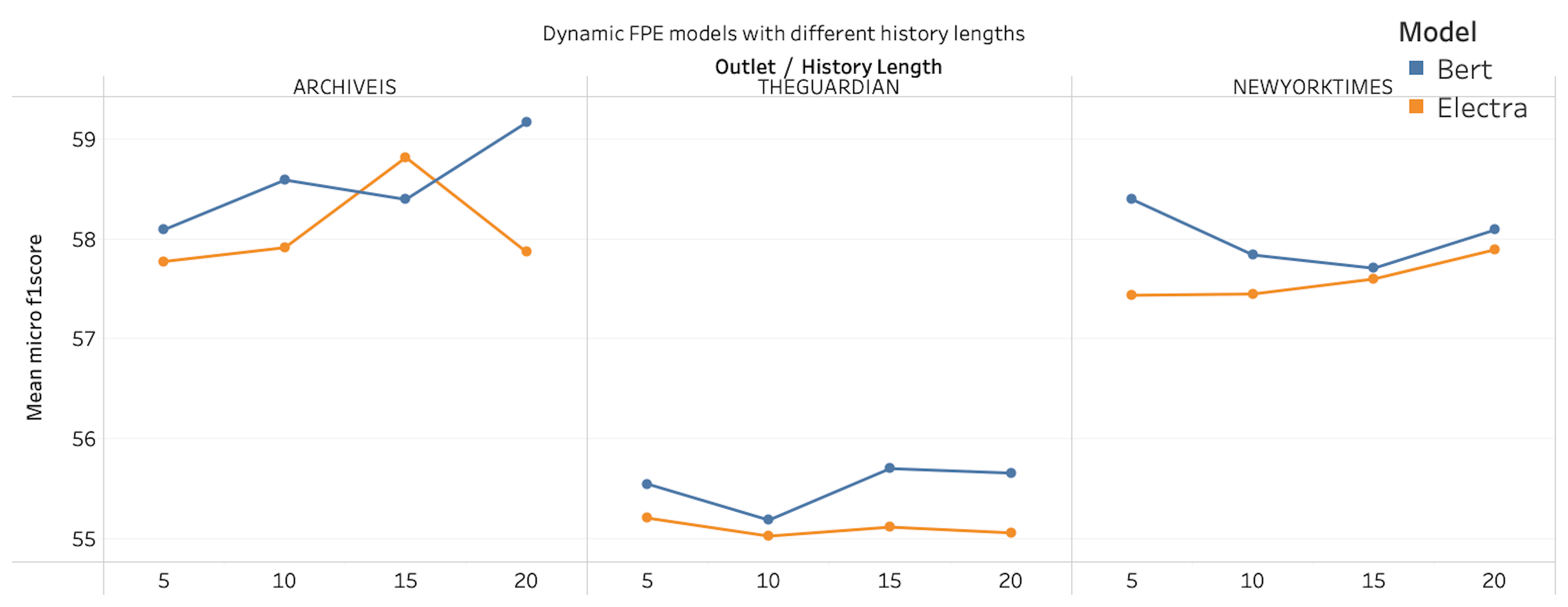}
    \caption{Impact of dynamic history length on micro F1-score in dynamic FPE model for different outlets with history length as x-axis and mean micro F1-score as y-axis. Overall, scores improve with increase in history length, especially for BERT(DistilBERT) models consistently. }
    \label{fig:dFPE_length}
\end{figure*}

\subsection{Implementation}

As discussed in Section \ref{sectionmodel}, we have primarily two models - \textbf{static FPE} and \textbf{dynamic FPE}- with different variants of each by using two types of contextual encoders- DistilBERT and ELECTRA, along with frozen and trained parameters. For static FPE models, we have taken an arbitrary history length of 12 article-comment pairs, so $s =$ 12 for fingerprinting. For dynamic FPE models, we have taken relevant dynamic history length of 15 article-comment pairs, $r =$ 15, after comparing micro F1-scores for various lengths from 5 to 20. We discuss about the impact of dynamic history length on performance in Section \ref{sec:results}. In all cases, a hidden layer of GRU with a dimension of 256 is created. We have trained all model variants for 10 epochs and saved the model based on the mean micro F1-score over all four sentiments on the validation dataset. Usually, the best model is achieved around 5-6 epochs. 

We use Adam optimizer with weight decay \citep{DBLP:conf/iclr/LoshchilovH19} and a schedule of learning rate (lr) that decreases following the values of the cosine function between the initial lr set in the optimizer to 0, with several hard restarts, after a warmup period during which it increases linearly between 0 and the initial lr set to 0.001 in the optimizer.  We implemented the model using PyTorch lightning, a wrapper for PyTorch. The code is released at \href{https://github.com/kishoret04/OpinionPrediction.git}{GitHub}.

\begin{table}[t]

\small
\begin{tabularx}{0.48\textwidth}{XXXXXXl}

 \hline
 
 \textbf{Model} &\textbf{Type} &\textbf{BERT (parameters)}  &\textbf{history type} &\textbf{history length}\\
 \hline
 sFPE-BERT-frozen	&static FPE &DistilBERT (frozen) &static &12\\ \hline
 sFPE-BERT-trained	 &static FPE &DistilBERT (trained) &static &12\\ \hline
 sFPE-ELEC-frozen	 &static FPE &ELECTRA (frozen) &static &12\\ \hline
 sFPE-ELEC-trained	 &static FPE &ELECTRA (trained) &static &12\\ \hline
 dFPE-ELEC-frozen-shist	&dynamic FPE &ELECTRA (frozen) &static &12\\ \hline
 dFPE-ELEC-frozen-dhist	&dynamic FPE &ELECTRA (frozen) &dynamic &15\\ \hline
 dFPE-BERT-frozen-dhist	&dynamic FPE &DistilBERT (frozen) &dynamic &15\\ \hline
 
 \end{tabularx}
 \caption{Notation of models along with their configuration details about type of FPE model, BERT variant, history type, length of history.}
\label{table:model_config}
\end{table}

\section{Results and discussion} \label{sec:results}

 Table \ref{table:model_config} describes the model notation and corresponding configuration used in the results. Figure \ref{fig:final_results} shows that except for vader sentiment, all our model variants outperform the FPE baseline. In the remaining three sentiments, dynamic-FPE-ELECTRA (frozen) model with either dynamic or static history outperformed the remaining variants and also FPE baseline. This could be because of multiple reasons, and we discuss them below.

Firstly, contextual embedding, in place of static BPE embedding in FPE, of articles and comments is a key factor behind the superior performance. Specifically, the contextual embedding of article history with target article (reading track) and with corresponding comment history (writing track) has enabled us to generate a better representation of the input text.

Secondly, the dynamic FPE model, unlike the static FPE model, creates reading and writing fingerprints through GRU networks separately before concatenating them for sentiment prediction. With an extra GRU hidden layer in the reading track compared to the writing track, we have been able to create a higher level of understanding of the temporal relationship between target article and history articles. 
From these fingerprints, we also found that users with the closest fingerprints in the euclidean space are found to have higher prediction accuracy than that of farther fingerprint users.

\begin{figure}
    \centering
    \includegraphics[scale=0.035]{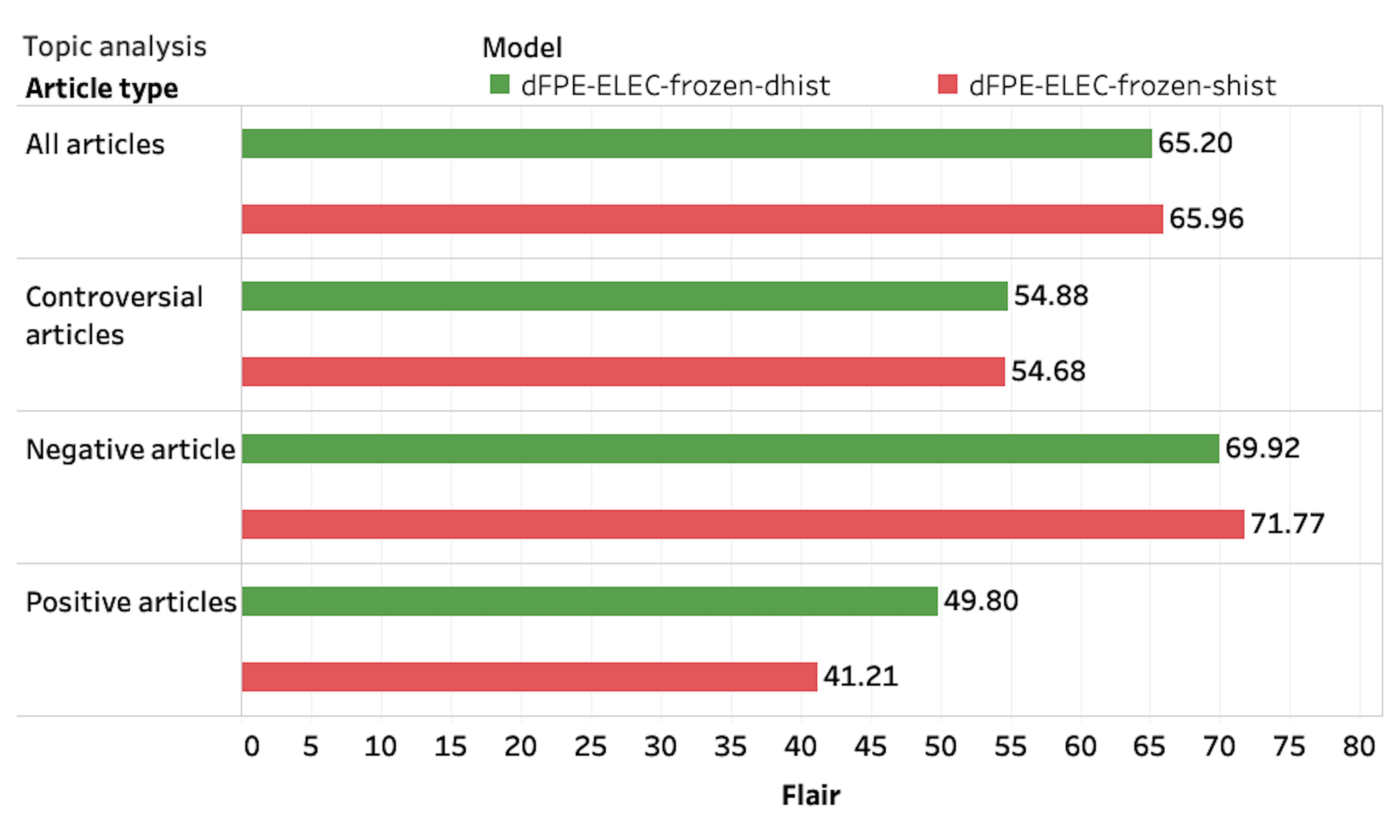}
    \caption{Impact of nature of article on Flair micro F1-score for dynamic FPE ELECTRA model with static and dynamic history for Archiveis outlet }
    \label{fig:Topic_analysis}
\end{figure}

Further, the ELECTRA-based model outperforms DistilBERT-based model in most of the cases, despite having only 20\% of the number of parameters of DistilBERT-based model. This reiterates the result of  \citealp{DBLP:journals/corr/abs-2003-10555} that novel pretraining by a discriminative model that predicts whether each token in the corrupted input was replaced by a generator sample or not performs better than masked language modeling pretraining method of DistilBERT, even for a small model.

From Figure \ref{fig:Topic_analysis}, we also observe that articles with a large proportion of comments with negative sentiment have a higher micro F1-score compared to controversial (equally positive and negative sentiment comments) and positive comments (predominant positive sentiment comments).

Other important inferences are: a) performance increases with length of relevant history for dynamic FPE models in general as shown in Figure \ref{fig:dFPE_length}; b) BERT variant models with frozen parameters have better performance than those with all parameters trained. This may be because the pretrained model is trained on a much larger corpus than our problem dataset and so it has better language understanding. 

\section{Limitations and future work}
According to \citealp{https://doi.org/10.1111/j.1541-1338.2011.00536.x}, fingerprinting for predicting behavior is second-generation biometrics, which is different from first-generation biometrics that uses characteristics that are visible to the naked eye such as facial images, hand fingerprints. In this context, user fingerprinting in our model can be loosely classified under behavior fingerprinting. We did linguistic opinion and content-based user fingerprinting as a response history embedding for a user. In this section, we briefly discuss the limitations of the dataset and model, and future research direction. 

Firstly, the newspaper articles, in general, may be biased in terms of story selection, tone, and organizing of the story. The users(readers) may also have an implicit bias - attitudes that unconsciously affect individual thoughts and actions- and confirmation bias - the tendency to support information that confirms their beliefs. To address these biases, we would like to extend our model to datasets that are not related to news articles.

Moreover, representations encoded in the models often inadvertently perpetuate undesirable social biases from the data on which they are trained. NLP models, especially neural embeddings, may perpetuate these biases towards race, religion, gender and disability \citep{DBLP:conf/acl/HutchinsonPDWZD20, DBLP:conf/naacl/ManziniLBT19, DBLP:conf/acl/SapCGCS19}. Though the BERT variant based sentence
encoders exhibit less bias than previous models \citep{DBLP:conf/naacl/MayWBBR19}, we would also like to experiment with other sentence encoders to measure the bias in our predictions in future work.

Another limitation of our approach is that we used only article titles rather than whole content. This would be more critical when the title is misleading, for instance in satirical articles. Moreover, we have not experimented with multilingual models of pretrained BERT variants.   

For future research, these experiments can be extended for whole article content and use various attention mechanisms to generate better fingerprints and also generate author profiles based on their reading history. Further, BERT variants trained in other languages can also be used.

\section{Conclusion}

In this paper, we propose a novel dynamic fingerprinting technique based on BERT variants and RNN networks to predict a user's sentiment to an unseen article based on reading-writing history. Two variants of our model extract relevant history in two different ways and create contextual embedding for articles read by a user conditioned with target article and also corresponding comments. Finally, we used RNN to interpret temporal relationship and create a fingerprint, which is used to predict unseen target article. Our models demonstrated state-of-the-art performance on a real-world dataset. From our experiments, we found that performance saturates after an optimum length of relevant history.

\section*{Acknowledgments}

Research was supported in part by grants NSF 1838147, NSF 1838145, ARO W911NF-20-1-0254. The views and conclusions contained in this document are those of the authors and not of the sponsors. The U.S. Government is authorized to reproduce and distribute reprints for
Government purposes notwithstanding any copyright notation herein. Finally, we are greatly appreciative of the anonymous reviewers for their time and constructive comments.

\bibliographystyle{acl_natbib}
\bibliography{OpinionPrediction}


\end{document}